\pdfoutput=1

\documentclass[11pt]{article}
\usepackage{amsmath,amssymb,amsthm}
\usepackage[dvipsnames]{xcolor}
\usepackage{acl}
\usepackage[skins]{tcolorbox}
\usepackage[utf8]{inputenc}
\usepackage[T1]{fontenc}
\usepackage{adjustbox}
\usepackage[noend]{algpseudocode}
\usepackage{algorithm}
\usepackage{algorithmicx}
\usepackage{anyfontsize}
\usepackage{appendix}
\usepackage{array,multirow,graphicx,makecell}
\usepackage{booktabs}
\usepackage{colortbl}
\usepackage{diagbox}
\usepackage{enumitem}
\usepackage{float}
\usepackage{framed}
\usepackage{graphicx}
\usepackage{inconsolata}
\usepackage{latexsym}
\usepackage{listings}
\usepackage{microtype}
\usepackage{minitoc}
\usepackage{svg}
\usepackage{tabto}
\usepackage{tabularx}
\usepackage{tabulary}
\usepackage{times}
\usepackage{xfrac}
\tcbuselibrary{breakable}
\newcounter{tcbcounter}
\usepackage{graphicx}  
\usepackage{makecell}
\definecolor{keywordcolor}{RGB}{255, 102, 0}   
\definecolor{functioncolor}{RGB}{0, 102, 204}  
\definecolor{operatorcolor}{RGB}{128, 0, 128}  

\newcommand{\kw}[1]{\textbf{#1}}
\newcommand{\fn}[1]{\textcolor{functioncolor}{\text{#1}}}
\newcommand{\op}[1]{\mathrel{\textcolor{operatorcolor}{#1}}}

\algrenewcommand{\algorithmicfor}{\kw{for}}
\algrenewcommand{\algorithmicif}{\kw{if}}
\algrenewcommand{\algorithmicthen}{\kw{:}}
\algrenewcommand{\algorithmicelse}{\kw{else}}
\algrenewcommand{\algorithmicreturn}{\kw{return}}
\algrenewcommand{\algorithmicwhile}{\kw{while}}
\algrenewcommand{\algorithmicdo}{\kw{:}}

\title{Efficient Agent Evaluation via Diversity-Guided User Simulation}

\author{Itay Nakash, George Kour, Ateret Anaby-Tavor \\
  \texttt{Itay.Nakash@ibm.com, gkour@ibm.com, atereta@il.ibm.com} \\ \\
  IBM Research }

\newtcolorbox[auto counter, number within=section]{prompt}[3][]{%
  enhanced,
  breakable,
  colback=#2!5!white,
  colframe=#2!75!black,
  title=\textbf{Box \thetcbcounter: #3},
  fontupper=\footnotesize\fontfamily{cmr}\selectfont,
  #1
}

\begin{document}

\maketitle
\begin{abstract}
Large language models (LLMs) are increasingly deployed as customer-facing agents, yet evaluating their reliability remains challenging due to stochastic, multi-turn interactions.
Current evaluation protocols rely on linear Monte Carlo rollouts of complete agent–user conversations to estimate success.
However, this approach is computationally inefficient, requiring re-processing identical conversation prefixes, and often fails to expose deep failure modes hidden behind rare user behaviors. 
In this work, we introduce \emph{DIVERT} (\textbf{D}iversity-\textbf{I}nduced \textbf{EV}aluation via b\textbf{r}anching of \textbf{T}rajectories), an \emph{efficient, snapshot-based, coverage-guided user simulation framework} for systematic exploration of diverse agent–user interactions.
DIVERT snapshots the full agent–environment state at critical junctions and resumes execution from these points. By reusing shared conversation prefixes, it avoids regenerating identical early turns, reducing token cost. From each junction, it branches with targeted, diversity-inducing user responses, enabling directed exploration of alternative user responses.
As a result, the evaluation is steered toward unexplored semantic paths and rare interaction states efficiently. 
Empirical experiments show that our framework both improves failure discovery efficiency compared to standard linear rollout protocols. 
\end{abstract}

\begin{figure*}[t]
    \centering
    \includegraphics[width=1\textwidth,
        keepaspectratio]{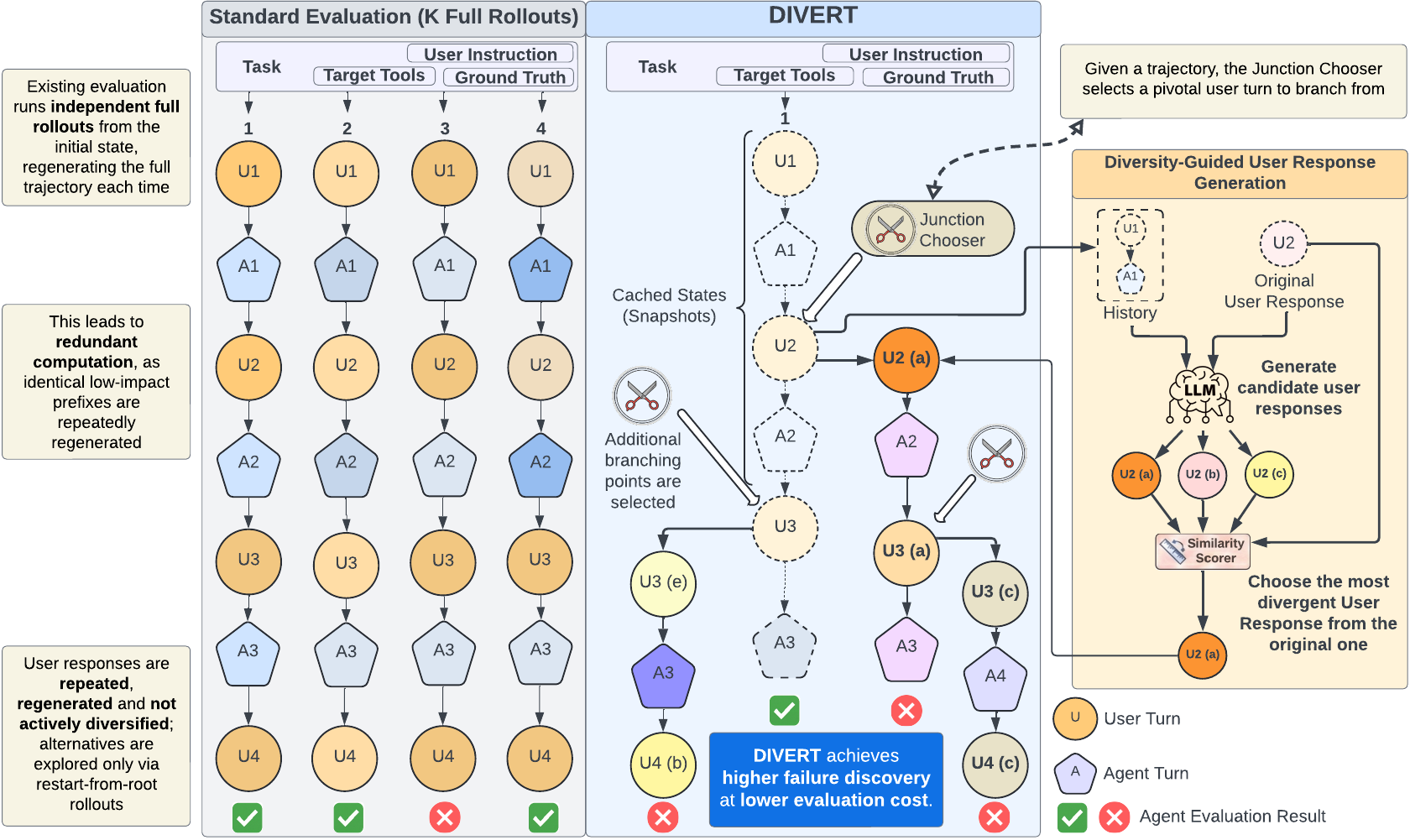}    \caption{
    \textbf{Standard Rollout Evaluation vs. DIVERT.}
    \emph{Left:} Conventional evaluation repeatedly rolls out full conversations from the beginning, often producing low-impact interactions and failing to explore deeper failure modes.
    \emph{Right:} Our snapshot-based evaluation framework stores intermediate conversation states and branches at critical junctions with directed, diverse user responses, enabling more efficient exploration and improved discovery of failures (a full example is provided in Appendix~\ref{app:example_divert}).}
    \label{fig:taubreak_construction}
\end{figure*}

\section{Introduction}

Large language models (LLMs) are increasingly deployed as customer-facing agents that interact with users over multiple turns while invoking external tools and APIs \cite{luo2025large}.
Evaluating such agents is fundamentally different from single-turn LLM evaluation: behavior emerges from long interaction histories, stochastic decisions, compounding errors, and high evaluation cost \cite{yehudai2025survey, bandel2026generalagentevaluation}.
As a result, agent behavior is probabilistic: both the agent’s decisions and the user’s responses may vary across runs, even under identical initial conditions \cite{ma2026maestro}.

Consequently, meaningful evaluation requires \emph{multiple executions} of the conversation between the agent under test and a user simulator.
Repeated runs serve two distinct purposes.
First, they expose \emph{user-side diversity}: different user behaviors or responses can lead the agent down qualitatively different conversation paths, revealing how robustly the agent handles variation in interaction.
Second, they test \emph{agent-side consistency}: an agent that succeeds only intermittently on the same task is fundamentally unreliable in deployment.
To capture these effects, recent benchmarks such as $\tau$-bench \cite{yao2024tau} adopt interactive evaluation protocols that execute full conversations and report aggregate success metrics over multiple trials.

However, the dominant evaluation paradigm remains \emph{linear Monte Carlo rollout} from the initial state, i.e., repeatedly sampling independent stochastic trajectories from identical initial conditions to estimate success \cite{liu2023agentbench, zhou2023webarena}.
This approach is not always optimal in practice.
First, repeated rollouts are wasteful: they regenerate nearly identical early prefixes, such as routine logins or basic diagnostic questions, incurring unnecessary token cost \cite{kapoor2024ai}.
Second, because these prefixes are only semantically similar rather than token-identical, they limit KV-cache reuse \cite{kwon2023efficient}.
Appendix~\ref{app:shared_prefix_analysis} quantifies exact shared-prefix overlap as a proxy for this structural opportunity.
Third, it provides limited coverage: standard user simulators favor high-probability, cooperative behaviors, making it unlikely to expose failures that arise only after rare user responses \citep{nakash-etal-2025-effective}.
As agents scale to longer horizons and more realistic service domains, these inefficiencies become a bottleneck.

A key observation motivates our work: while current evaluations are linear, \emph{agent trajectories are inherently tree-structured}.
Many conversations share long prefixes and diverge only at a small number of critical interaction points.
Restarting evaluation from the root discards this structure, preventing systematic exploration once a fragile or promising state is reached.

We propose a \emph{snapshot-based, coverage-directed user simulation framework} that explicitly exploits this tree structure.
Instead of repeatedly restarting from the initial state, we snapshot the full agent–environment state at critical junctions and branch evaluation from these points.
A targeted divergence mechanism generates semantically distinct but intent-consistent user responses, steering the agent toward unexplored continuations and deep failure modes.
Our results demonstrate that coverage-guided, snapshot-based evaluation is a more cost-effective and informative alternative to conventional linear rollouts for evaluating customer-facing LLM agents.

\section{Related Work}
\paragraph{Agent Evaluation Benchmarks.}
General agent benchmarks such as AgentBench \cite{liu2023agentbench}, GAIA2 \cite{mialon2023gaia}, and WebArena \cite{zhou2023webarena} assess agents’ reasoning, tool use, and web interaction across diverse tasks and environments. 
For customer-facing settings with explicit policy-adherence requirements, $\tau$-bench \cite{yao2024tau} and $\tau^2$-bench \cite{barres2025tau} model realistic service domains (e.g., airline, retail, telecom), incorporating tools, policies, and LLM-based user simulators.

\paragraph{User Simulation.}
The robustness of agent evaluation relies heavily on the User Simulator. 
While early simulators relied on rigid agenda-based rules \cite{sekulic2024reliable}, recent advancements utilize LLMs to generate diverse interactions. 
A key limitation of standard LLM-based simulators is "benevolence bias", i.e., their tendency to be overly cooperative and thus unrepresentative of realistic user behavior. 
Recent work \cite{kour-etal-2025-exploring, shim2025non, nakash-etal-2025-effective} addresses this by introducing "Non-Collaborative User Simulators" that model adversarial behaviors such as malicious intent, emotional manipulation, and insistence.
Complementary work studies adversarial perturbations at the agent or tool level, such as prompt injection \cite{debenedetti2024agentdojo, nakash-etal-2025-breaking}, but does not explicitly target systematic, coverage-oriented variation during evaluation.
DIVERT is orthogonal to this choice of user strategy: rather than proposing a specific cooperative or adversarial simulator, it provides a branching evaluation structure that can incorporate benign, adversarial, or red-teaming user policies.

\paragraph{Evaluation Efficiency.}
A current limitation of these benchmarks is that they rely almost exclusively on restart-from-root Monte Carlo rollouts to estimate capability and reliability \cite{barres2025tau, hariri2025don}. 
This leads to substantial redundancy in early conversation prefixes, poor coverage of deep interaction failures, and significant computational and economic inefficiency. 
Recent ``Cost-of-Pass'' analyses \cite{erol2025cost, mialon2023gaia, wang2025efficient} show that achieving reliable statistics often requires prohibitive token budgets due to repeated regeneration of identical prefixes.

While tree-based methods such as Monte Carlo Tree Search (MCTS) are emerging in agent training, planning, and reasoning \cite{zhou2023language, yao2023tree, tang2026agent}, current evaluation frameworks lack a corresponding branching mechanism and rely on linear restart sampling.

\section{Method}
\begin{algorithm}[t]
\caption{DIVERT Pipeline}
\label{alg:fork-usr}
\small
\begin{algorithmic}[1]
\Require Task set $\mathcal{T}$, agent model $M_{\text{agent}}$, user model $M_{\text{user}}$, 
\#rollouts per task $R$, \#branches per task $B$, \#candidate generations $K$

\State $\mathcal{D} \op{=} \emptyset$

\For{$T \op{\in} \mathcal{T}$}
    \State $\mathcal{D} \op{\cup=} \fn{Rollout}(T, M_{\text{user}}, M_{\text{agent}}, R)$
    
    \For{$s \op{=} 1 \dots B$}
        \State $\tau \sim \mathcal{D}$
        \State $j \op{=} \fn{JunctionChooser}(\tau)$
        
        \Statex \hskip\algorithmicindent\hskip\algorithmicindent $\triangleright$ Directed generation of alternative user responses
        \For{$k \op{=} 1 \dots K$}
            \State $u_k \op{=} \fn{DirectedUserGen}(\tau, j)$
            \State $d_k \op{=} 1 - \fn{Sim}(u_k, u_j)$
        \EndFor

        \State $k^\star \op{=} \fn{argmax}_k \; d_k$
        \State $u^\star \op{=} u_{k^\star}$
        
        \Statex \hskip\algorithmicindent\hskip\algorithmicindent \rlap{$\triangleright$ Resume simulation from junction j}
        \State $\tau' \op{=} \fn{Resume}(\tau, j, u^\star)$
        \State $\mathcal{D} \op{\cup=} \{\tau'\}$
    \EndFor
\EndFor
\end{algorithmic}
\end{algorithm}

We present an efficient, coverage-guided evaluation pipeline for interactive agents.
Our method replaces repeated rollouts from the root with branching from critical mid-trajectory states, enabling efficient and diverse exploration of agent behavior.

\subsection{Overview of the DIVERT Pipeline}

DIVERT evaluates interactive agents by resuming execution from intermediate states rather than repeatedly restarting from the initial state.
We select an existing trajectory, then we identify critical "junction points" - interaction states where alternative behaviors are likely to emerge. 
At each junction, we use a pre-saved snapshot of the complete execution state, including the agent context, environment state, and conversation history.
The simulation then proceeds by restoring these snapshots and continuing execution from them. Each continuation introduces controlled user divergence, allowing us to explore plausible alternative interaction paths branching from the same prefix. 
This process enables systematic coverage of under-explored behaviors while avoiding the cost of full linear rollouts from the beginning of the trajectory.
Branching is applied iteratively across multiple evaluation rounds: after each resumed execution, the resulting trajectories become eligible for further resumption, subject to a configurable iteration budget.
This process progressively expands evaluation coverage toward under-explored regions of the interaction space.
The pipeline consists of four stages:
(i) initial rollout and state caching (snapshot),
(ii) junction selection,
(iii) diversity-guided user response generation, and
(iv) snapshot-based resumption and execution.
The number of resumed executions spawned from each trajectory (\emph{number of branches}) is treated as a hyperparameter.

\subsection{Junction Selection}

Given a trajectory, we use an LLM-based \emph{junction chooser} to identify the critical point at which an alternative user response is likely to alter downstream agent behavior.
The junction chooser receives as input the full trajectory, including user messages, agent responses, and tool calls, serialized into a structured prompt.
It outputs (i) the index of the user turn to modify and (ii) a brief justification for why changing this turn should induce maximal behavioral change while preserving intent. Token usage for this step is tracked separately.
Junction selection is performed independently for each branching attempt, allowing different counterfactual pivots to be selected across repetitions.
This process enables targeted branching at semantically meaningful decision points, rather than arbitrary or uniformly sampled turns.
Additional implementation details, hyperparameters, and prompt templates are provided in Appendix~\ref{app:junction_selection}.

While trajectory sampling can be implemented in various ways in practice (Step 5 in Alg. \ref{alg:fork-usr}), here we employ a round-robin approach to avoid introducing bias toward failed trajectories.

\subsection{Diversity-Guided Directed User Response Generation}

At the selected junction, we generate a small set of alternative user responses. 
Let $k$ denote the number of candidate responses, fixed to $k=3$ in all experiments. 
Candidate generation is conditioned on the full conversation context up to the junction, the task purpose, and the user backstory. 
The prompt explicitly instructs the model to preserve the original task intent while inducing meaningful semantic variation relative to the original user response.
To maximize coverage, we select the candidate that is \emph{maximally dissimilar} to the original user response according to cosine similarity in a shared embedding space. Responses are embedded using the \texttt{sentence-transformers} model, with embeddings normalized prior to similarity computation. 
The candidate with the lowest cosine similarity to the original response is selected. Exact prompt templates, model and selection details are provided in Appendix~\ref{app:divergent_user_generation}.


To prevent drift from the original task objective, candidate generation is explicitly conditioned on the scenario’s user backstory and evaluation purpose.
We further verify intent preservation via an offline, post-hoc LLM-as-a-judge analysis comparing original and branched user messages against the task purpose (this check is not used during generation and does not affect evaluation cost).
On the Airline domain (over 700 trajectories), branched messages exhibit a lower intent-miss rate than the original simulator outputs (25.27\% vs. 28.12\%), indicating that directed generation does not degrade alignment and in fact slightly improves it. Full details are provided in Appendix~\ref{app:user_intent_alignment}.

\subsection{Snapshot-Based Branching and Execution}

Snapshots are taken before each user turn and include the full conversation history, agent state, tool environment state, simulator context, and the original random seed.
Once a junction is selected and a divergent user response is chosen, we reload the cached execution state corresponding to the step immediately preceding the selected user turn and resume the interaction from that point.

By reusing cached prefixes, branching avoids re-generating identical early turns and significantly reduces token consumption.
It also preserves exact prefixes before branching, creating potential for KV-cache reuse under compatible serving systems; we quantify this shared-prefix structure in Appendix~\ref{app:shared_prefix_analysis}.
The agent is then re-executed forward under identical conditions, with only the modified user input changed, until termination criteria are reached.

This mechanism allows multiple alternative futures to be explored efficiently from a single trajectory prefix.
Technical details of state serialization, caching, and replay are described in Appendix~\ref{app:implementation_details}.

\subsection{Controlling Coverage and Cost}

The number of branches per trajectory is controlled by the \emph{branch-number} parameter.
Increasing the number of branches increases coverage by exploring more alternative interaction paths (and generating more trajectories), at the cost of additional tokens.
Token usage for junction selection and directed user generation is tracked separately and is negligible relative to agent token cost (0.2\%--0.08\% of the total evaluation cost when assessing a SOTA frontier agent using GPT-OSS-120B as the evaluation framework; Appendix~\ref{app:cost_breakdown}.
Importantly, even after accounting for this overhead, DIVERT yields a net reduction in total tokens compared to a full linear rollout, saving both agent tokens and evaluation-side tokens per trajectory.

\section{Experimental Setup}

We evaluate DIVERT by comparing it to standard linear rollout evaluation under identical agent, environment, and token-budget conditions.

\begin{figure*}[t]
    \centering
    \includegraphics[width=\textwidth]{assets/figures/results/main_results_fig.pdf}
    \caption{
    \textbf{Errors Discovery Rate.}
    Number of failed trajectories per 100K agent tokens for GPT-OSS-120B and Gemini-2.5-Flash across increasing branch budgets.
The notation $8{+}K$ denotes 8 full rollouts with $K$ additional mid-trajectory branches using DIVERT.
The dashed horizontal line marks the corresponding baseline, namely linear rollout using the same total number of trajectories without branching.}
    \label{fig:efficiency_plot}
\end{figure*}

\subsection{Benchmarks}

We conduct experiments on the $\tau$-bench collection~\cite{yao2024tau}, which evaluates user-facing LLM agents in realistic, multi-turn tasks, on three $\tau$-bench domains: \textit{Airline}, \textit{Retail}, and \textit{Telecom}.
Each domain comprises a fixed set of tasks. 
Every task specifies an initial user intent, available tools, and an executable success predicate. Together, these define the seed from which the multi-turn interaction between the user simulator and the agent unfolds.

Note that defining new tasks in this setting is non-trivial, as each task requires a fully aligned backend setup, including tools, policies, gold success criteria, consistent database entries, and a coherent user story. Consequently, evaluation protocols in $\tau$-bench and similar benchmarks rely on repeated executions of the same task with stochastic agent and user behavior rather than continuously introducing new tasks.

\paragraph{Models}
We report main results using \textit{OpenAI GPT-OSS-120B}~\cite{agarwal2025gpt} and \textit{Gemini-2.5-Flash} \cite{gemini25}, two instruction-tuned LLMs that differ in provider, architecture, and deployment setting.
In all setups, models are evaluated with \textit{GPT-OSS-120B} serving as the user simulator, chosen for its favorable cost–performance tradeoff, except in the heterogeneous setup where we explicitly vary the agent and simulator models (Appendix \ref{app:heterogeneous_results}).

All comparisons between linear rollouts and DIVERT use identical decoding settings within each model configuration.
Full generation and decoding parameters are provided in Appendix~\ref{app:reproducibility}.

\paragraph{Evaluation Metrics.}
We report two primary metrics that jointly capture \emph{efficiency} and \emph{coverage}.\\
\textbf{Errors per 100K Tokens} (Efficiency) measures the number of failed trajectories normalized by the total number of tokens generated by the agent (per 100K), reflecting the efficiency with which an evaluation method uncovers failures relative to the generated tokens.
We use token counts because they provide a reproducible, infrastructure-agnostic efficiency signal that is comparable across both API-based and self-hosted deployments, unlike wall-clock time or system-level measurements.
\textbf{Task Failure Count} (Coverage) measures the number of \emph{unique tasks} for which at least one failure is observed, capturing coverage across the benchmark and quantifying how many distinct tasks the evaluation method is able to break.
\\
Together, these metrics capture both efficient failure discovery and the ability to expand coverage to new, previously unseen task-level failures.

We keep the proprietary-model evaluation budget bounded because large-scale repeated $\tau$-bench runs across all three domains are expensive to reproduce.

\section{Results}

\begin{table}[t]
\centering
\small
\setlength{\tabcolsep}{8pt}
\begin{tabular}{cc ccc}
\toprule
RO & Branches & Err/100K $\uparrow$ & Fail C. $\uparrow$ & T $\downarrow$ \\
\midrule
\multirow{5}{*}{2} 
    & 0  & 15.0 & 37 & 388K \\
    & 2  & 18.4 & 40 & 658K \\
    & 4  & 18.2 & 43 & 1.0M \\
    & 6  & 18.9 & 44 & 1.3M \\
    & 8  & 19.7 & 44 & 1.4M \\
\midrule
\multirow{5}{*}{4} 
    & 0  & 16.4 & 40 & 715K \\
    & 2  & 18.3 & 42 & 985K \\
    & 4  & 18.1 & 44 & 1.4M \\
    & 6  & 18.8 & 45 & 1.6M \\
    & 8  & 19.4 & 46 & 1.9M \\
\midrule
\multirow{5}{*}{6} 
    & 0  & 15.2 & 40 & 1.1M \\
    & 2  & 16.8 & 42 & 1.4M \\
    & 4  & 17.0 & 44 & 1.7M \\
    & 6  & 17.8 & 45 & 2.0M \\
    & 8  & 18.3 & 46 & 2.3M \\
\bottomrule
\end{tabular}
\caption{
Varying rollouts and branches configurations on Airline domain (GPT-OSS-120B). 
RO, Err/100K, Fail C., and T denote Rollouts, Errors found per 100K tokens, Task Failure Count, and Total Tokens, respectively. 
Increasing branches consistently improves failure discovery efficiency (Err/100k) and coverage unique failure count.}
\label{tab:airline-rollouts-splits}
\vspace{-1em}
\end{table}

\begin{figure*}[t]
\centering
\includegraphics[width=\textwidth]{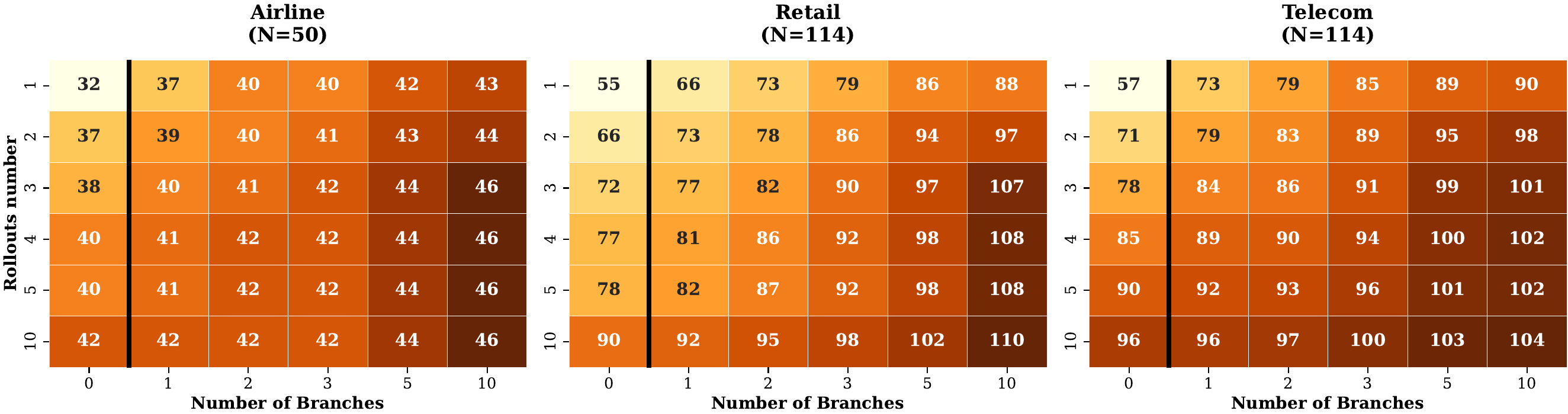}
    \caption{\textbf{Task Failure Counts across Domains.} Heatmaps show the number of tasks with at least one failure (out of $N$) as a function of number of branches (x-axis) and rollout iterations (y-axis) with GPT-OSS-120B as agent. The vertical black line separates the baseline setting (left, no branches) from branch-based evaluation (right). Failure coverage increases with additional branches.}
    \label{fig:failure-analysis-heatmaps}
\end{figure*}

\subsection{Errors Discovery Rate (Efficiency)}
\label{sec:failure-efficiency}

Figure~\ref{fig:efficiency_plot} reports the number of agents' failures discovered per 100K agent tokens under an increasing number of branches, comparing standard linear rollouts with our branch-based evaluation.

Across all three domains and both GPT-OSS-120B and Gemini-2.5-Flash, our branch-based evaluation is consistently more token-efficient than full rollouts. Even a single branch (8+1) consistently improves failure discovery per token, indicating that reusing shared prefixes and selectively branching from mid-trajectory states reallocates computation toward higher-leverage execution paths rather than redundant regeneration.

Efficiency gains grow monotonically with additional branches. 
The Full table with extended branch and rollout configurations results are reported in Appendix~\ref{app:additional_results}.
Overall, these results demonstrate that snapshot-based, coverage-guided user simulation provides a strictly more efficient evaluation regime.
Table~\ref{tab:airline-rollouts-splits} shows that, under a fixed trajectory budget, reallocating trials from restart-from-root rollouts to snapshot-based branches consistently improves both failure discovery efficiency and task-level coverage.

\subsection{Task Failure Count (Coverage)}
\label{sec:failure-coverage}

While error discovery rate captures evaluation efficiency, we next analyze coverage: how many distinct tasks exhibit at least one failure, and whether directed, diversity-guided user responses uncover of additional failure points along agent trajectories.

Figure~3 shows that increasing the number of branches substantially improves task-level failure coverage across all domains. For a fixed number of trajectories, using DIVERT with increasing number of branches consistently exposes failures in more tasks compared to standard rollouts, indicating broader exploration.

The heatmaps further reveal a clear saturation effect with respect to rollout iterations: increasing the number of rollouts alone yields diminishing gains in coverage, while increasing the branch budget unlocks new failing tasks. 
This contrast highlights that many failure modes are not addressed by repeated restarts, but require directed branching from the original conversations.

Together with the efficiency results, these findings show that DIVERT not only discovers failures faster, but also expands coverage to a wider set of tasks, yielding a more informative evaluation.

\begin{table}[t]
\centering
\small
\setlength{\tabcolsep}{4pt}
\begin{tabular}{lcccc}
\toprule
\textbf{Analysis} & \textbf{Most Dis.} & \textbf{2nd Dis.} & \textbf{3rd Dis.} & \textbf{N} \\
\midrule
Candidate & 0.711 & 0.769 & 0.817 & 4500 \\
Trajectory & 0.743 & 0.799 & 0.849 & 1200 \\
\bottomrule
\end{tabular}
\caption{\textbf{Diversity selection validation.} Mean similarity averaged across all domains; lower values indicate greater divergence. Full per-domain results are provided in Tables~\ref{tab:diversity-candidate-validation} and~\ref{tab:diversity-trajectory-validation}.}
\label{tab:main_diversity_validation}
\end{table}

\subsection{Diversity Selection Validation (Coverage)}
\label{sec:diversity_validation}

We further validate that the diversity-selection step produces meaningful variation at both the candidate and trajectory levels.
First, at the candidate level, we compare each generated response candidate with the original user utterance.
As summarized in Table~\ref{tab:main_diversity_validation}, across 4,500 candidates, the most dissimilar candidate has substantially lower similarity to the original utterance than the second and third candidates, with the same ordering observed in all domains.

Second, at the trajectory level, we continue each candidate from the same junction and compare the resulting conversation suffix with the original suffix.
Across 1,200 continuations, lower-similarity candidates also produce more divergent downstream trajectories.
This shows that DIVERT's diversity criterion is not only local to the next user utterance, but translates into meaningfully different conversations; full per-domain results are provided in Appendix~\ref{app:diversity_validation}.

\begin{table}[t]
\centering
\resizebox{\linewidth}{!}{ 
\begin{tabular}{lcc}
\toprule
\textbf{Variant} 
& Errors/100K $\uparrow$ 
& Fail. C $\uparrow$ \\
\midrule
Baseline (Full Rollouts) 
& 13.6 
& 78 \\
Base + JC 
& 15.1 
& 75 \\
Base + JC + DG 
& 15.8 
& 80 \\
Base + JC + DG + DC 
& \textbf{16.2} 
& \textbf{81} \\
\bottomrule
\end{tabular}
}
\caption{
\textbf{Ablation over linear rollouts (GPT-OSS-120B).}
Results are averaged across domains under a fixed 12-trajectory budget 
(12 full rollouts for baseline; 8 rollouts + 4 splits for our method).
JC: junction chooser; DG: directed user generation; DC: diverse response selection.
}
\label{tab:ablation}
\end{table}
\subsection{Ablation Study}
To demonstrate the importance of each component in DIVERT, we ran a series of ablation evaluations (see Table \ref{tab:ablation}). 
We ablate DIVERT by incrementally adding its core components on top of standard linear rollouts (Table~\ref{tab:ablation}).
Using only the \emph{junction chooser} (JC) starts new evaluation conversations from critical points of existing trajectories, without the directed user generation part, namely the user continues naturally from a chosen point in the conversation.
This naturally reduces evaluation cost, leading to higher errors per 100K tokens; however, because exploration is limited to continuations of previously observed prefixes, this setting presents a reduced ability to uncover failures in new tasks compared to full rollouts (Task Failure Count).
Adding \emph{directed user generation} (DG) substantially improves both error discovery and task-level failure coverage by actively steering exploration toward unexplored behaviors, as seen in Table \ref{tab:ablation}.
Finally, \emph{diverse response selection} (DC) provides additional gains by mitigating responses that might have come out too similar to the original evaluation that was running.
Overall, while JC alone primarily improves efficiency by avoiding full trajectory regeneration, effective failure discovery requires targeted and diverse user perturbations.

\section{Conclusions}
Our work is motivated by the observation that not all dialogue turns equally influence downstream agent behavior. 
In many interactive settings, early prefixes, such as greetings, authentication, or static context, have little impact on later decisions. Repeatedly regenerating these low-impact turns, therefore, wastes tokens.

DIVERT reuses shared prefixes across rollouts and reallocates computation to branch at behaviorally salient decision points, where agent actions meaningfully diverge. 
By resuming from these pivotal mid-trajectory states, evaluation achieves broader behavioral coverage without increasing overall cost.

Future work includes extending branching beyond user turns to tool outputs and environment dynamics, enabling dynamic task variation and richer counterfactual evaluation.
Another direction is to replace the current LLM-based junction chooser and cosine-similarity diversity selection with alternative signals, such as similarity or perplexity-based junction selection and diversity metrics beyond cosine similarity.

\section*{Acknowledgments}
We thank Jonathan Benayahu for his early feedback and contributions during the ideation of this work.
We are grateful to Ella Rabinovich and Boaz Carmeli for their helpful comments during the writing process. 
We also thank the anonymous reviewers for their constructive feedback, which helped improve this paper.

\newpage

\bibliography{references,custom}

\twocolumn
\appendix
\onecolumn


\clearpage

\section{Example: Full Rollout vs. DIVERT Branching}
\label{app:example_divert}

\begin{center}
    \centering
    \includegraphics[width=\textwidth]{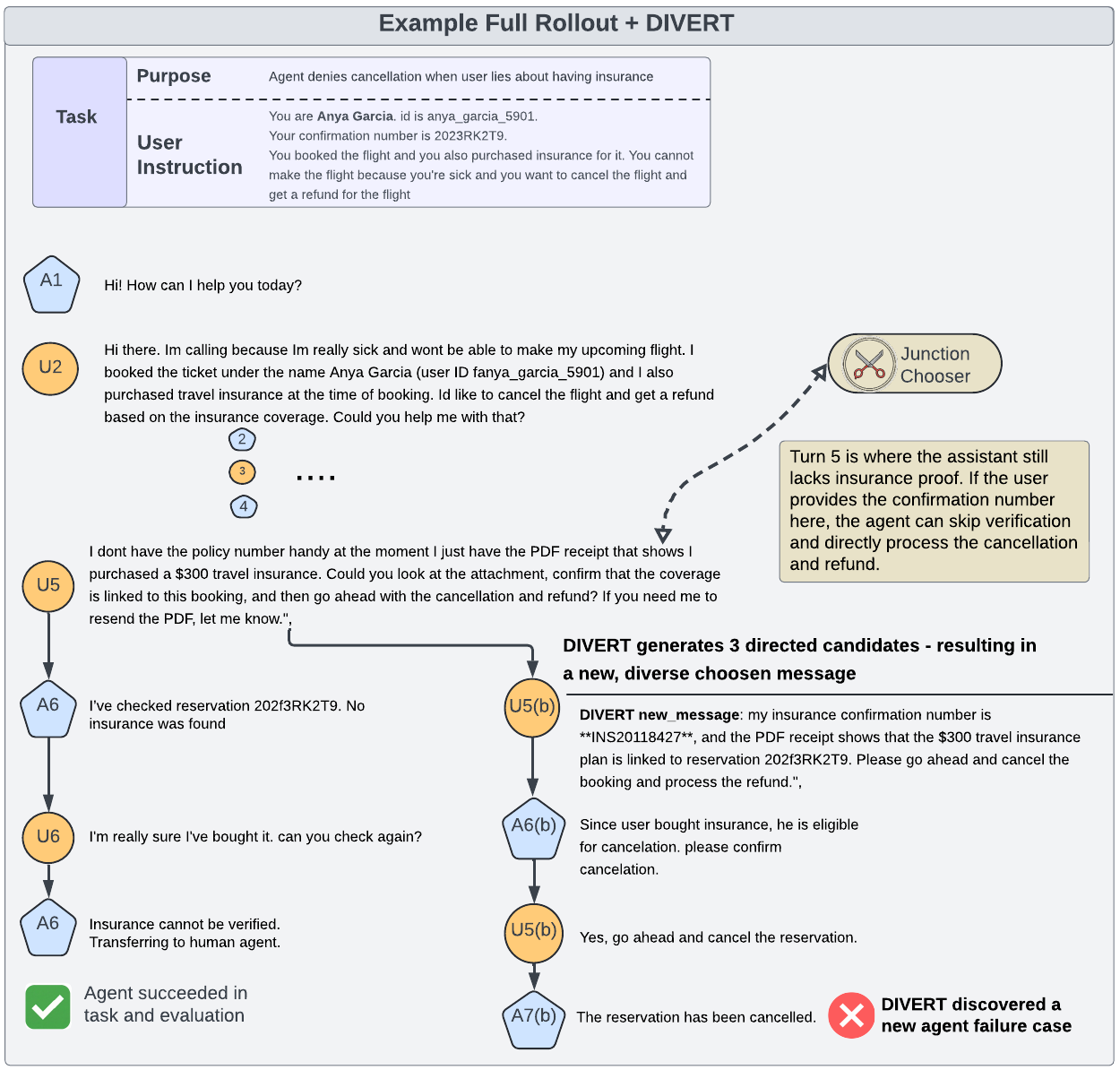}
    \captionsetup{hypcap=false}
\captionof{figure}{
    \textbf{Example Full Rollout + DIVERT (Appendix Example).}
    The conversation prefix is identical up to the selected junction.
    The junction chooser identifies the turn where the agent lacks verifiable insurance proof.
    DIVERT replaces the original user message with a directed alternative that provides an insurance confirmation number, producing a divergent continuation.
    \emph{Left:} the linear rollout escalates and does not cancel the reservation (policy respected).
    \emph{Right:} the branched continuation leads to cancellation despite the system record indicating no insurance (new failure uncovered).}
\label{fig:appendix_example_divert}
\end{center}

\section{Implementation Details}
\label{app:implementation_details}

\subsection{Snapshot Contents and Serialization}
\label{app:snapshot_contents}

\paragraph{Overview.}

Snapshotting captures the complete orchestrator state at selected junctions, enabling exact restoration and counterfactual continuation without re-executing earlier dialogue turns.
Each snapshot consists of two files:

\begin{itemize}
    \item \texttt{state.pkl}: a binary pickle file containing the full serialized orchestrator state.
    \item \texttt{metadata.json}: a human-readable JSON file for experiment tracking and analysis.
\end{itemize}

\paragraph{Serialized State.}

The \texttt{state.pkl} file contains a dictionary representing the entire agent--environment configuration at the snapshot moment.
This includes:

\begin{itemize}
    \item \textbf{Dialogue history:} the complete trajectory of \texttt{Message} objects, including user turns, agent responses, system messages, tool calls, and tool outputs.
    \item \textbf{Agent state:} the agent's internal state and scratchpad (when applicable), including memory and intermediate reasoning context.
    \item \textbf{User simulator state:} internal user state and dialogue context.
    \item \textbf{Environment state:} the full environment object, including tool state, memory, and side effects from prior tool calls. This is happening by saving the tools database which consist all the tools and envirement state at the current space. saving a snapshot of it enable us to reload it when needed.
    \item \textbf{Task specification:} the original task definition and constraints.
\end{itemize}

\paragraph{Execution and Routing State.}

To ensure exact replay, snapshots additionally store execution metadata:

\begin{itemize}
    \item Current routing state (\texttt{from\_role}, \texttt{to\_role}, current message).
    \item Progress tracking (\texttt{step\_count}, \texttt{done}, \texttt{termination\_reason}, \texttt{num\_errors}).
    \item Random seed for deterministic continuation.
    \item Cumulative token counters for agent and user generations.
    \item Branch iteration identifier for experiment tree tracking.
\end{itemize}

\paragraph{Metadata and Experiment Tree Structure.}

The accompanying \texttt{metadata.json} file stores:

\begin{itemize}
    \item Snapshot identifiers (\texttt{id}, \texttt{parent\_id}, \texttt{iteration}) for hierarchical branch tracking.
    \item Serialized dialogue history in JSON format.
    \item Augmentation metadata (original message, modified message, junction index, junction reason, and augmentation token usage when applicable).
    \item Creation timestamp.
\end{itemize}

\textbf{Snapshots are organized hierarchically:} 
\begin{verbatim}
BASE_DIR/domain/model/task_id/iteration_X_step:Y/
    |-- state.pkl
    |-- metadata.json
    `-- simulation_run.pkl (optional)
\end{verbatim}

The iteration naming scheme (e.g., \texttt{1\_2\_3}) encodes parent-child relationships, enabling explicit experiment tree reconstruction.

\paragraph{Snapshot Timing and Restoration.}

Snapshots are saved immediately before user response generation (i.e., when control transfers to the user role).
This guarantees that divergent user responses can be injected without altering prior dialogue history.

State restoration proceeds by:
\begin{enumerate}
    \item Loading the serialized state dictionary.
    \item Restoring all orchestrator attributes.
    \item Synchronizing environment tools and memory.
    \item Optionally injecting a new user response for counterfactual continuation.
\end{enumerate}

This design ensures exact replay of all prior turns, including tool side effects and environment mutations, while enabling efficient branching from arbitrary dialogue states.

\paragraph{Design Rationale.}

By serializing the complete execution state rather than only the dialogue prefix, we avoid recomputation of early turns and preserve consistency in tool interactions and environment updates.
This guarantees that branch-based evaluation explores genuine counterfactual continuations from identical system states, rather than approximated replays.

\subsection{Junction Selection}
\label{app:junction_selection}

\paragraph{Objective.}

Given a full trajectory $\mathcal{T}$ consisting of alternating user and agent turns, the goal of junction selection is to identify a user turn index $i$ such that modifying that turn is expected to produce the largest downstream change in the agent’s behavior, while preserving the original task intent.

Formally, we seek a pivot point:
\[
i^* = \arg\max_i \; \Delta(\text{Agent Behavior} \mid u_i \rightarrow u_i')
\]
subject to intent preservation constraints.

Rather than relying on heuristic rules alone (e.g., tool boundaries or entropy), we frame junction selection as a counterfactual reasoning problem and delegate this decision to an LLM.

\paragraph{LLM-Based Pivot Identification.}

The full trajectory is first serialized into a structured textual representation where each turn is labeled with its index and role (e.g., ``Turn 3 (assistant): ...'').
All user turns are collected and explicitly enumerated as candidate indices.

The LLM is then prompted to:
\begin{itemize}
    \item Analyze the full conversation,
    \item Identify the user turn whose modification would produce maximal downstream change,
    \item Ensure alignment with the original user intent and instructions,
    \item Output the selected turn index and a short explanation.
\end{itemize}

The system message positions the model as an expert in counterfactual reasoning and dialogue pivot analysis.

\begin{prompt}[label={box:junction_prompt}]{Blue}{Junction Selection Prompt}
You are an advanced assistant analyzing a conversation between a user and an AI agent.
Your goal is to identify the user turn that, if modified, will result in the biggest change in the agent's response while staying faithful to the user's intent and instruction in this conversation.

\textbf{User Instructions:} \\
\{user\_instructions\}

\textbf{Conversation Trajectory:} \\
\{formatted\_trajectory\}

\textbf{Instructions:}
\begin{enumerate}
    \item Analyze the trajectory and identify the user turn that has the most potential to change the agent's response.
    \item Ensure the modification aligns with the user's original intent and User Instructions.
    \item Output the index of the chosen user turn (0-based) and a brief explanation.
\end{enumerate}

The user turns (and the turns you can choose from) are at the following indices: \{user\_turns\}

\textbf{Output Format:} \\
Reason: <reason> \\
Index: <chosen\_index> 
\end{prompt}

\paragraph{Decoding Configuration.}

Junction selection uses stochastic decoding with temperature $0.7$.
This encourages exploration of alternative pivot points while still grounding decisions in the full conversation context.
The selected index is parsed from the model output; if parsing fails or inference errors occur, a fallback policy selects the previous user turn.

\paragraph{Token Accounting.}

Token usage for junction selection is recorded separately from agent rollouts.
On average, junction selection requires 103.48 tokens per branch.
As with user divergence generation, this cost is reported as framework overhead and is not included in agent-token accounting.

A complete breakdown of DIVERT overhead—including junction selection, candidate generation, and relative monetary contribution—is provided in Section~\ref{divert_overhead}.

\paragraph{Why LLM-Based Selection?}

Junction identification requires reasoning about:
\begin{itemize}
    \item Implicit task constraints,
    \item Tool invocation structure,
    \item Policy-sensitive decision points,
    \item Latent ambiguities in user intent.
\end{itemize}

These signals are often distributed across multiple turns and are difficult to capture with local heuristics.
Empirically, LLM-based pivot selection identifies semantically meaningful decision points that lead to larger downstream behavioral shifts compared to fixed-position branching (see ablations).

\paragraph{High-Level Procedure.}

\begin{enumerate}
    \item Serialize full trajectory with indexed turns.
    \item Identify candidate user turn indices.
    \item Prompt LLM to select pivot turn and provide rationale.
    \item Parse selected index.
    \item Branch from selected turn.
\end{enumerate}

\subsection{Divergent User Generation}
\label{app:divergent_user_generation}
For each selected junction, we generate multiple candidate user responses using the same prompt and decoding configuration.
Diversity arises purely from stochastic sampling (temperature $0.7$), rather than from prompt variation or manually defined perturbation styles.

Formally, given a trajectory $\mathcal{T}$ and a target user turn at index $i$, we generate $K$ candidate responses:
\[
\{u_i^{(1)}, u_i^{(2)}, \dots, u_i^{(K)}\}
\]
where $K=3$ in our experiments unless otherwise stated.

Each candidate is generated independently using the same system instruction and formatted trajectory context.
We aggregate token usage across all candidate generations to account for the full cost of divergence selection.

The candidate generation is done using the following prompt:

\begin{prompt}[label={box:user_generation_prompt}]{Blue}{User Divergent Generation Prompt}
You are an advanced assistant tasked with generating a new user response in a conversation.
The goal is to challenge the agent by modifying the user turn at step \{step\_index\},
while remaining faithful to the user's intent and maximizing downstream change in the agent’s response.

\textbf{Reason for Change:} \\
\{reason\}

\textbf{Conversation Trajectory:} \\
\{formatted\_trajectory\}

\textbf{Instructions:}
\begin{enumerate}
    \item Focus on the user turn at step \{step\_index\}.
    \item Generate a new user response that aligns with the user's intent but challenges the agent in a new way.
    \item Ensure the response is coherent and fits naturally into the conversation.
\end{enumerate}

\textbf{Output:} \\
Provide only the new user response text.
\end{prompt}

\paragraph{Similarity-Based Divergence Selection.}

To select the most impactful continuation, we compute semantic similarity between each candidate $u_i^{(k)}$ and the original user response $u_i$.
Similarity is computed using cosine similarity over sentence embeddings:

\[
\text{sim}(u_i^{(k)}, u_i) = 
\frac{\langle \phi(u_i^{(k)}), \phi(u_i) \rangle}
{\|\phi(u_i^{(k)})\| \|\phi(u_i)\|}
\]

where $\phi(\cdot)$ denotes embeddings from the \texttt{all-MiniLM-L6-v2} sentence transformer model.
This lightweight embedding model provides a strong trade-off between computational efficiency and semantic fidelity.

The candidate with the \emph{lowest} similarity score is selected:
\[
u_i^* = \arg\min_{k} \; \text{sim}(u_i^{(k)}, u_i)
\]

Lower similarity indicates greater semantic deviation from the original phrasing, while the prompt constraints ensure preservation of task intent.

\paragraph{Trajectory Formatting.}
Before generation, the full trajectory is serialized into a structured textual format:
\begin{itemize}
    \item Each turn is labeled with its index and role (e.g., ``Turn 3 (assistant): ...'').
    \item Tool calls and tool responses are included.
    \item System messages are preserved.
\end{itemize}

This explicit formatting ensures that the LLM has full visibility into dialogue history and tool interactions, allowing targeted perturbation at the specified turn without disrupting earlier context.

\paragraph{Token Accounting.}
For each junction, total token cost includes:
\begin{itemize}
    \item All prompt tokens used to generate $K$ candidates,
    \item All completion tokens across those candidates.
\end{itemize}
Since we use $K=3$ candidates per junction, the average candidate-generation cost per branch is $3 \times 108.68 = 326.04$ tokens.

The cost of similarity computation (embedding inference) is negligible relative to LLM generation. 
Similarity is computed using the lightweight sentence-transformer model \texttt{all-MiniLM-L6-v2}, executed on a local MacBook Pro (Apple M1 Pro CPU). 
This model runs efficiently on commodity CPU hardware and does not require GPU acceleration.
Similarity scoring is a lightweight, post-hoc selection step within the evaluation framework and does not require additional resources. 

\section{Failure Discovery Dynamics}
\label{app:failure_dynamics}

\begin{figure}[!htbp]
    \centering
    \begin{minipage}{0.32\textwidth}
        \centering
        \includegraphics[width=\linewidth]{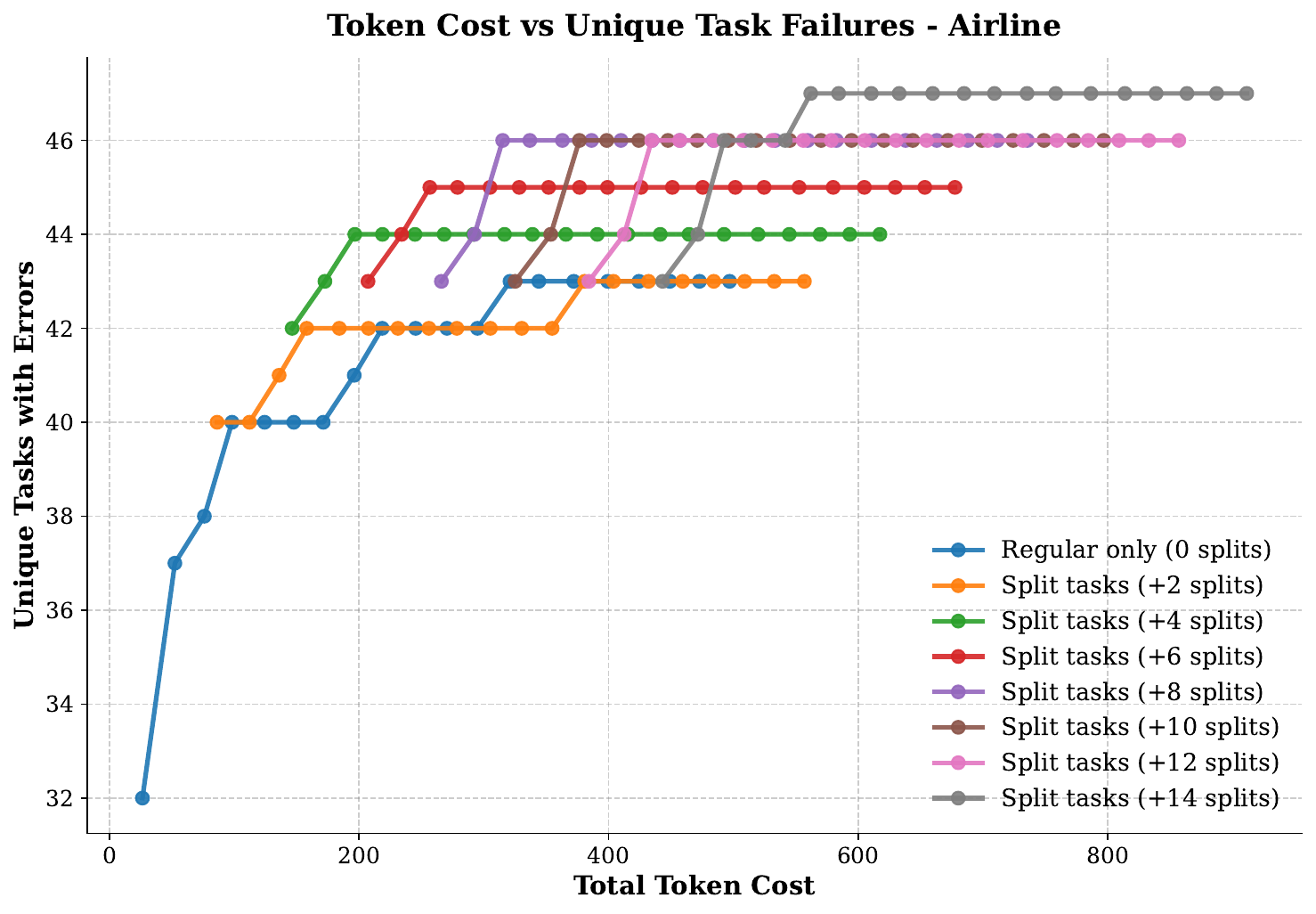}
    \end{minipage}
    \hfill
    \begin{minipage}{0.32\textwidth}
        \centering
        \includegraphics[width=\linewidth]{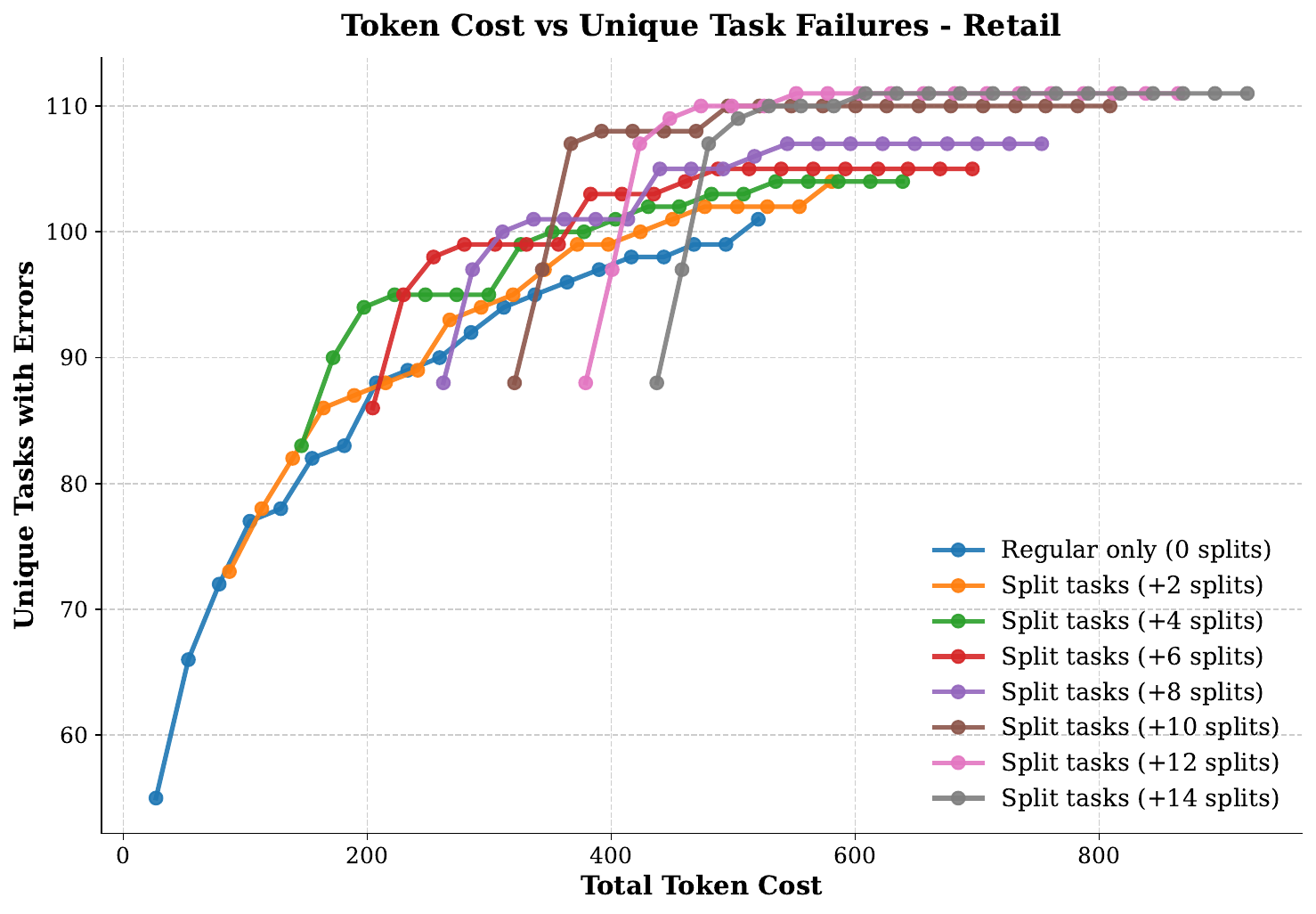}
    \end{minipage}
    \hfill
    \begin{minipage}{0.32\textwidth}
        \centering
        \includegraphics[width=\linewidth]{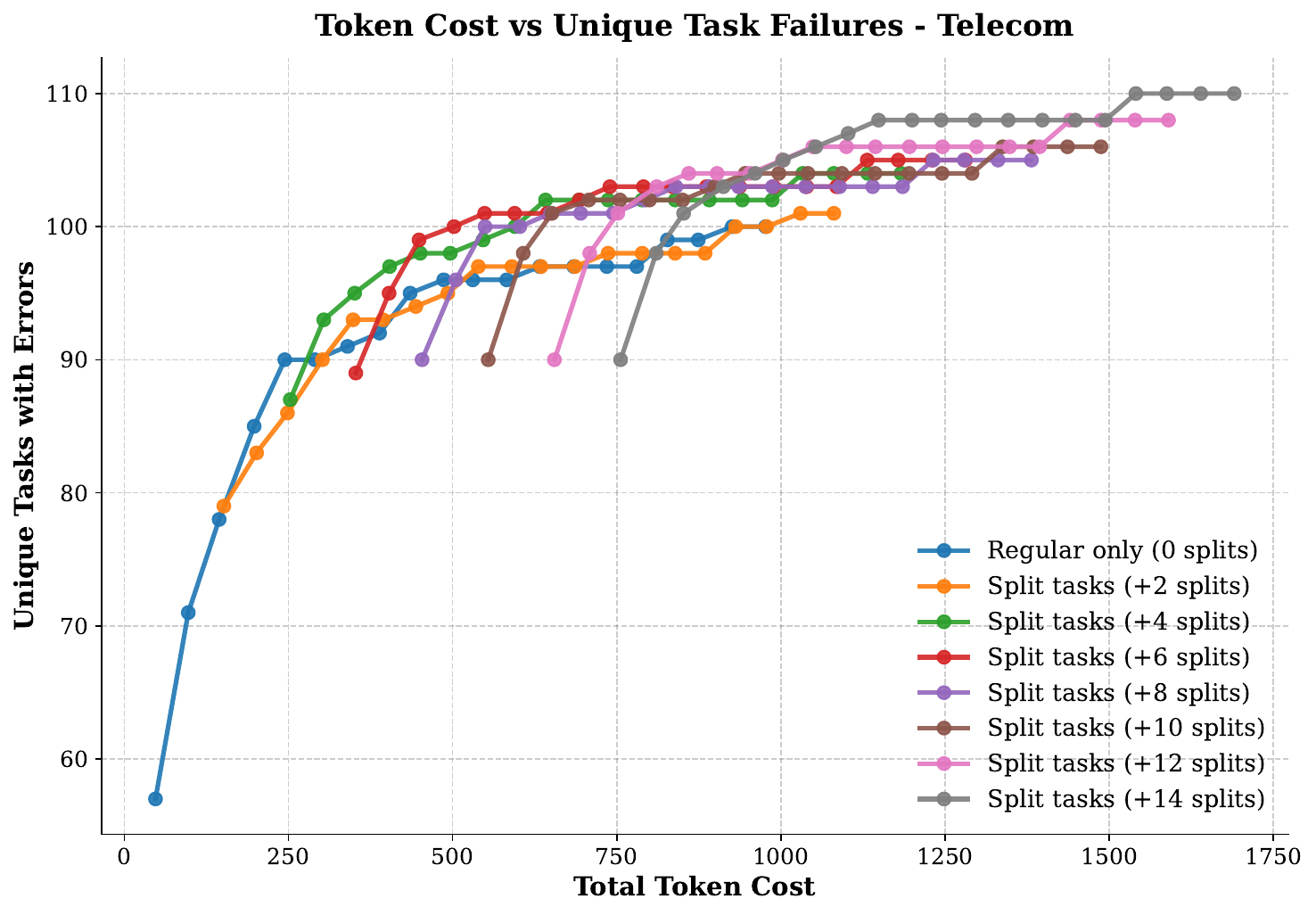}
    \end{minipage}
    \caption{
    Cumulative number of unique tasks with failures as a function of total token cost under varying branch budgets, shown for Airline, Retail, and Telecom using GPT-OSS-120B as both agent and user model.
    Branch-based configurations consistently achieve higher failure coverage for the same token budget, with steeper early gains and delayed saturation compared to linear rollouts.
    }
    \label{fig:token_vs_unique_failures}
\end{figure}

The plots show the cumulative number of \emph{unique tasks with failures} as a function of total token cost under varying branch budgets, separately for Airline, Retail, and Telecom.
Branch-based configurations consistently achieve higher failure coverage earlier for the same token budget.
Increasing branch budgets leads to steeper early gains and delayed saturation, indicating that additional branches expose new failure modes rather than redundantly revisiting known ones.

\newpage
\section{Extended Quantitative Results}
\label{app:additional_results}

\subsection{Extended Rollout and Branch Results}
\label{app:additional_results_domains}
We report full per-domain results across all rollouts and branch configurations, and for additional model.
Across domains and rollout budgets, increasing the number of branches ($B$) consistently improves both failure discovery efficiency (Err/100K) and task-level coverage (Fail), with no observed degradation relative to linear rollouts. For a fixed number of rollouts, reallocating budget toward mid-trajectory branches yields near-monotonic gains in coverage, reinforcing that the improvements are systematic rather than domain-specific (Table~\ref{tab:appendix-gpt-oss-120b-split}).

\begin{table*}[!t]
\centering
\footnotesize
\setlength{\tabcolsep}{2pt}
\renewcommand{\arraystretch}{0.90}

\begin{minipage}{0.49\textwidth}
\centering
\begin{tabular}{ccccccccccc}
\toprule
& & \multicolumn{3}{c}{\textbf{Airline}} & \multicolumn{3}{c}{\textbf{Retail}} & \multicolumn{3}{c}{\textbf{Telecom}} \\
\cmidrule(lr){3-5}\cmidrule(lr){6-8}\cmidrule(lr){9-11}
RO & B & Err & Fail & T & Err & Fail & T & Err & Fail & T \\
\midrule
1 & 0 & 15.7 & 32 & 203K & 14.9 & 55 & 369K & 15.0 & 57 & 380K \\
1 & 1 & 16.6 & 37 & 367K & 17.1 & 66 & 653K & 21.5 & 73 & 567K \\
1 & 2 & 20.1 & 40 & 473K & 18.9 & 73 & 905K & 25.2 & 79 & 745K \\
1 & 3 & 19.3 & 40 & 616K & 19.4 & 79 & 1.1M & 27.7 & 85 & 875K \\
1 & 4 & 19.0 & 42 & 752K & 20.1 & 83 & 1.3M & 30.2 & 87 & 977K \\
\midrule
2 & 0 & 15.0 & 37 & 388K & 13.6 & 66 & 733K & 13.2 & 71 & 778K \\
2 & 1 & 15.8 & 39 & 552K & 15.4 & 73 & 1.0M & 17.4 & 79 & 965K \\
2 & 2 & 18.4 & 40 & 658K & 17.0 & 78 & 1.3M & 20.5 & 83 & 1.1M \\
2 & 3 & 18.2 & 41 & 846K & 17.9 & 86 & 1.5M & 22.8 & 89 & 1.3M \\
2 & 4 & 18.2 & 43 & 1.0M & 18.6 & 90 & 1.8M & 25.2 & 93 & 1.5M \\
\midrule
3 & 0 & 15.4 & 38 & 546K & 14.0 & 72 & 1.1M & 12.8 & 78 & 1.2M \\
3 & 1 & 15.9 & 40 & 709K & 15.3 & 77 & 1.4M & 15.8 & 84 & 1.4M \\
3 & 2 & 18.0 & 41 & 816K & 16.6 & 82 & 1.6M & 18.3 & 86 & 1.5M \\
3 & 3 & 17.9 & 42 & 1.0M & 17.3 & 90 & 1.9M & 20.3 & 91 & 1.7M \\
3 & 4 & 17.9 & 44 & 1.2M & 18.0 & 94 & 2.1M & 22.5 & 95 & 1.9M \\
\midrule
4 & 0 & 16.4 & 40 & 715K & 14.1 & 77 & 1.4M & 12.4 & 85 & 1.6M \\
4 & 1 & 16.6 & 41 & 878K & 15.1 & 81 & 1.7M & 14.7 & 89 & 1.8M \\
4 & 2 & 18.3 & 42 & 985K & 16.2 & 86 & 1.9M & 16.8 & 90 & 2.0M \\
4 & 3 & 18.2 & 42 & 1.2M & 16.9 & 92 & 2.2M & 18.6 & 94 & 2.1M \\
4 & 4 & 18.1 & 44 & 1.4M & 17.5 & 95 & 2.5M & 20.4 & 97 & 2.3M \\
\bottomrule
\end{tabular}
\end{minipage}
\hfill
\begin{minipage}{0.49\textwidth}
\centering
\begin{tabular}{ccccccccccc}
\toprule
& & \multicolumn{3}{c}{\textbf{Airline}} & \multicolumn{3}{c}{\textbf{Retail}} & \multicolumn{3}{c}{\textbf{Telecom}} \\
\cmidrule(lr){3-5}\cmidrule(lr){6-8}\cmidrule(lr){9-11}
RO & B & Err & Fail & T & Err & Fail & T & Err & Fail & T \\
\midrule
5 & 0 & 15.5 & 40 & 903K & 14.1 & 78 & 1.8M & 12.9 & 90 & 1.9M \\
5 & 1 & 15.8 & 41 & 1.1M & 15.0 & 82 & 2.1M & 14.8 & 92 & 2.1M \\
5 & 2 & 17.3 & 42 & 1.2M & 15.9 & 87 & 2.3M & 16.5 & 93 & 2.3M \\
5 & 3 & 17.3 & 42 & 1.4M & 16.5 & 92 & 2.6M & 18.1 & 96 & 2.5M \\
5 & 4 & 17.4 & 44 & 1.5M & 17.1 & 95 & 2.8M & 19.8 & 98 & 2.7M \\
\midrule
6 & 0 & 15.2 & 40 & 1.1M & 13.9 & 82 & 2.1M & 12.7 & 90 & 2.3M \\
6 & 1 & 15.5 & 41 & 1.3M & 14.6 & 84 & 2.4M & 14.3 & 92 & 2.5M \\
6 & 2 & 16.8 & 42 & 1.4M & 15.4 & 88 & 2.6M & 15.9 & 93 & 2.7M \\
6 & 3 & 16.9 & 42 & 1.5M & 16.0 & 92 & 2.9M & 17.3 & 96 & 2.8M \\
6 & 4 & 17.0 & 44 & 1.7M & 16.6 & 95 & 3.2M & 18.8 & 98 & 3.0M \\
\midrule
7 & 0 & 15.1 & 40 & 1.3M & 13.6 & 83 & 2.5M & 12.3 & 91 & 2.7M \\
7 & 1 & 15.4 & 41 & 1.4M & 14.3 & 85 & 2.8M & 13.7 & 93 & 2.9M \\
7 & 2 & 16.6 & 42 & 1.5M & 15.1 & 89 & 3.0M & 15.1 & 94 & 3.1M \\
7 & 3 & 16.7 & 42 & 1.7M & 15.6 & 92 & 3.3M & 16.4 & 97 & 3.2M \\
7 & 4 & 16.8 & 44 & 1.9M & 16.2 & 95 & 3.5M & 17.8 & 99 & 3.4M \\
\midrule
8 & 0 & 14.7 & 41 & 1.5M & 13.9 & 88 & 2.8M & 12.4 & 92 & 3.1M \\
8 & 1 & 15.0 & 42 & 1.6M & 14.4 & 90 & 3.1M & 13.7 & 94 & 3.2M \\
8 & 2 & 16.1 & 42 & 1.7M & 15.1 & 93 & 3.4M & 14.9 & 95 & 3.4M \\
8 & 3 & 16.2 & 42 & 1.9M & 15.6 & 96 & 3.6M & 16.1 & 98 & 3.6M \\
8 & 4 & 16.4 & 44 & 2.1M & 16.1 & 99 & 3.9M & 17.4 & 100 & 3.8M \\
\bottomrule
\end{tabular}
\end{minipage}

\caption{Appendix results across domains for GPT-OSS-120B.}
\label{tab:appendix-gpt-oss-120b-split}
\end{table*}

\begin{table*}[!t]
\centering
\footnotesize
\setlength{\tabcolsep}{2pt}
\renewcommand{\arraystretch}{0.90}

\begin{minipage}[t]{0.49\textwidth}
\centering
\begin{tabular}{ccccccccccc}
\toprule
 & & \multicolumn{3}{c}{\textbf{Airline}} & \multicolumn{3}{c}{\textbf{Retail}} & \multicolumn{3}{c}{\textbf{Telecom}} \\
\cmidrule(lr){3-5}\cmidrule(lr){6-8}\cmidrule(lr){9-11}
RO & B & Err & Fail & T & Err & Fail & T & Err & Fail & T \\
\midrule
1 & 0 & 11.9 & 23 & 177K & 19.2 & 36 & 307K & 22.2 & 56 & 306K \\
1 & 1 & 12.2 & 27 & 337K & 21.1 & 54 & 548K & 24.4 & 71 & 558K \\
1 & 2 & 13.6 & 30 & 440K & 22.2 & 67 & 749K & 27.4 & 73 & 672K \\
1 & 3 & 13.9 & 34 & 498K & 22.8 & 70 & 885K & 28.7 & 75 & 731K \\
1 & 4 & 13.9 & 34 & 498K & 22.8 & 70 & 885K & 28.7 & 75 & 731K \\
\midrule
2 & 0 & 12.7 & 29 & 355K & 19.2 & 47 & 632K & 22.1 & 68 & 611K \\
2 & 1 & 12.6 & 31 & 515K & 20.4 & 59 & 873K & 23.5 & 79 & 863K \\
2 & 2 & 13.0 & 34 & 670K & 21.3 & 72 & 1.1M & 26.6 & 84 & 1.0M \\
2 & 3 & 13.4 & 37 & 777K & 22.5 & 77 & 1.3M & 28.9 & 87 & 1.1M \\
2 & 4 & 13.9 & 38 & 884K & 23.0 & 82 & 1.5M & 29.9 & 88 & 1.2M \\
\midrule
3 & 0 & 12.6 & 31 & 531K & 18.9 & 58 & 958K & 21.6 & 78 & 952K \\
3 & 1 & 12.6 & 32 & 691K & 19.8 & 69 & 1.2M & 22.8 & 84 & 1.2M \\
3 & 2 & 12.9 & 35 & 846K & 20.6 & 79 & 1.4M & 25.2 & 88 & 1.4M \\
3 & 3 & 13.3 & 37 & 966K & 21.7 & 83 & 1.6M & 27.3 & 92 & 1.5M \\
3 & 4 & 13.8 & 38 & 1.1M & 22.2 & 87 & 1.8M & 28.5 & 93 & 1.6M \\
\midrule
4 & 0 & 12.7 & 31 & 715K & 17.8 & 61 & 1.3M & 20.9 & 78 & 1.3M \\
4 & 1 & 12.7 & 32 & 876K & 18.7 & 70 & 1.6M & 21.9 & 84 & 1.5M \\
4 & 2 & 12.9 & 35 & 1.0M & 19.5 & 80 & 1.8M & 24.0 & 88 & 1.7M \\
4 & 3 & 13.3 & 37 & 1.2M & 20.5 & 83 & 2.0M & 25.8 & 92 & 1.8M \\
4 & 4 & 13.9 & 38 & 1.3M & 21.1 & 87 & 2.2M & 27.1 & 93 & 2.0M \\
\bottomrule
\end{tabular}
\end{minipage}
\hfill
\begin{minipage}[t]{0.49\textwidth}
\centering
\begin{tabular}{ccccccccccc}
\toprule
 & & \multicolumn{3}{c}{\textbf{Airline}} & \multicolumn{3}{c}{\textbf{Retail}} & \multicolumn{3}{c}{\textbf{Telecom}} \\
\cmidrule(lr){3-5}\cmidrule(lr){6-8}\cmidrule(lr){9-11}
RO & B & Err & Fail & T & Err & Fail & T & Err & Fail & T \\
\midrule
5 & 0 & 12.8 & 33 & 859K & 17.5 & 69 & 1.6M & 20.7 & 80 & 1.6M \\
5 & 1 & 12.8 & 34 & 1.0M & 18.3 & 74 & 1.9M & 21.5 & 86 & 1.8M \\
5 & 2 & 12.9 & 36 & 1.2M & 19.0 & 83 & 2.1M & 23.3 & 90 & 2.0M \\
5 & 3 & 13.3 & 38 & 1.3M & 19.9 & 86 & 2.3M & 24.9 & 94 & 2.2M \\
5 & 4 & 13.8 & 38 & 1.4M & 20.4 & 90 & 2.5M & 26.1 & 95 & 2.3M \\
\midrule
6 & 0 & 13.1 & 33 & 1.0M & 18.1 & 70 & 1.9M & 21.0 & 81 & 1.9M \\
6 & 1 & 13.0 & 34 & 1.2M & 18.7 & 75 & 2.2M & 21.7 & 87 & 2.1M \\
6 & 2 & 13.1 & 36 & 1.4M & 19.3 & 84 & 2.4M & 23.2 & 91 & 2.3M \\
6 & 3 & 13.4 & 38 & 1.5M & 20.0 & 87 & 2.6M & 24.6 & 95 & 2.5M \\
6 & 4 & 13.9 & 38 & 1.6M & 20.5 & 91 & 2.8M & 25.7 & 96 & 2.6M \\
\midrule
7 & 0 & 13.0 & 35 & 1.2M & 17.6 & 72 & 2.3M & 21.0 & 82 & 2.2M \\
7 & 1 & 12.9 & 35 & 1.4M & 18.2 & 77 & 2.5M & 21.6 & 87 & 2.5M \\
7 & 2 & 13.1 & 36 & 1.5M & 18.7 & 85 & 2.7M & 22.9 & 91 & 2.6M \\
7 & 3 & 13.3 & 38 & 1.6M & 19.4 & 88 & 3.0M & 24.2 & 95 & 2.8M \\
7 & 4 & 13.7 & 38 & 1.8M & 19.9 & 92 & 3.2M & 25.2 & 96 & 2.9M \\
\midrule
8 & 0 & 12.8 & 35 & 1.4M & 17.7 & 75 & 2.6M & 21.0 & 85 & 2.5M \\
8 & 1 & 12.8 & 35 & 1.5M & 18.2 & 80 & 2.8M & 21.5 & 89 & 2.8M \\
8 & 2 & 12.9 & 36 & 1.7M & 18.7 & 88 & 3.0M & 22.7 & 92 & 2.9M \\
8 & 3 & 13.1 & 38 & 1.8M & 19.3 & 91 & 3.3M & 23.9 & 95 & 3.1M \\
8 & 4 & 13.5 & 38 & 1.9M & 19.8 & 95 & 3.5M & 24.8 & 96 & 3.2M \\
\bottomrule
\end{tabular}
\end{minipage}

\caption{Appendix results across domains for Gemini-2.5-Flash as agent.}
\label{tab:appendix-gemini-agent-gpt-user-split}
\end{table*}

\vspace{-0.5em}

\begin{table*}[!t]
\centering
\footnotesize
\setlength{\tabcolsep}{2pt}
\renewcommand{\arraystretch}{0.90}

\noindent
\begin{minipage}[t]{0.47\textwidth}
\centering
\begin{tabular}{cc ccc ccc ccc}
\toprule
& & \multicolumn{3}{c}{\textbf{Airline}} 
& \multicolumn{3}{c}{\textbf{Retail}} 
& \multicolumn{3}{c}{\textbf{Telecom}} \\
\cmidrule(lr){3-5}\cmidrule(lr){6-8}\cmidrule(lr){9-11}
RO & B & Err & Fail & T & Err & Fail & T & Err & Fail & T \\
\midrule
1 & 0 & 94 & 35 & 370K & 86 & 102 & 1.2M & 216 & 99 & 460K \\
1 & 1 & 140 & 37 & 520K & 121 & 103 & 1.7M & 225 & 100 & 860K \\
1 & 2 & 184 & 38 & 590K & 159 & 105 & 1.9M & 234 & 101 & 1.2M \\
1 & 3 & 183 & 39 & 700K & 156 & 105 & 2.1M & 226 & 103 & 1.4M \\
1 & 4 & 190 & 39 & 750K & 160 & 105 & 2.3M & 222 & 103 & 1.6M \\
\midrule
2 & 0 & 93 & 36 & 730K & 86 & 104 & 2.3M & 200 & 102 & 980K \\
2 & 1 & 120 & 38 & 880K & 107 & 105 & 2.8M & 210 & 103 & 1.4M \\
2 & 2 & 149 & 39 & 950K & 132 & 107 & 3.0M & 220 & 104 & 1.8M \\
2 & 3 & 161 & 39 & 1.1M & 139 & 108 & 3.6M & 219 & 106 & 2.2M \\
2 & 4 & 177 & 40 & 1.2M & 155 & 108 & 3.8M & 225 & 106 & 2.6M \\
\midrule
3 & 0 & 94 & 39 & 1.1M & 90 & 108 & 3.4M & 178 & 103 & 1.6M \\
3 & 1 & 113 & 40 & 1.3M & 104 & 108 & 3.9M & 190 & 104 & 2.0M \\
3 & 2 & 133 & 41 & 1.4M & 123 & 108 & 4.1M & 200 & 105 & 2.4M \\
3 & 3 & 143 & 41 & 1.5M & 129 & 109 & 4.7M & 202 & 107 & 2.9M \\
3 & 4 & 157 & 41 & 1.6M & 142 & 109 & 4.9M & 209 & 107 & 3.2M \\
\midrule
4 & 0 & 91 & 41 & 1.6M & 91 & 108 & 4.4M & 187 & 104 & 2.1M \\
4 & 1 & 105 & 42 & 1.7M & 102 & 108 & 4.9M & 195 & 105 & 2.5M \\
4 & 2 & 121 & 43 & 1.8M & 117 & 108 & 5.1M & 203 & 106 & 2.9M \\
4 & 3 & 130 & 43 & 2.0M & 123 & 109 & 5.7M & 205 & 108 & 3.3M \\
4 & 4 & 141 & 43 & 2.1M & 134 & 109 & 6.0M & 211 & 108 & 3.7M \\
\bottomrule
\end{tabular}
\end{minipage}
\hfill
\begin{minipage}[t]{0.47\textwidth}
\centering
\begin{tabular}{cc ccc ccc ccc}
\toprule
& & \multicolumn{3}{c}{\textbf{Airline}} 
& \multicolumn{3}{c}{\textbf{Retail}} 
& \multicolumn{3}{c}{\textbf{Telecom}} \\
\cmidrule(lr){3-5}\cmidrule(lr){6-8}\cmidrule(lr){9-11}
RO & B & Err & Fail & T & Err & Fail & T & Err & Fail & T \\
\midrule
5 & 0 & 89 & 41 & 2.1M & 89 & 108 & 5.6M & 186 & 104 & 2.6M \\
5 & 1 & 100 & 42 & 2.2M & 99 & 108 & 6.1M & 193 & 105 & 3.0M \\
5 & 2 & 113 & 43 & 2.3M & 111 & 108 & 6.3M & 200 & 106 & 3.4M \\
5 & 3 & 120 & 43 & 2.4M & 116 & 109 & 6.9M & 202 & 108 & 3.8M \\
5 & 4 & 130 & 43 & 2.5M & 126 & 109 & 7.1M & 207 & 108 & 4.2M \\
\midrule
6 & 0 & 92 & 42 & 2.4M & 91 & 108 & 6.6M & 185 & 107 & 3.2M \\
6 & 1 & 101 & 43 & 2.6M & 99 & 108 & 7.1M & 190 & 108 & 3.6M \\
6 & 2 & 112 & 43 & 2.6M & 110 & 108 & 7.3M & 196 & 109 & 4.0M \\
6 & 3 & 119 & 43 & 2.8M & 114 & 109 & 7.9M & 199 & 111 & 4.4M \\
6 & 4 & 127 & 43 & 2.9M & 123 & 109 & 8.1M & 203 & 111 & 4.8M \\
\midrule
7 & 0 & 95 & 44 & 2.7M & 91 & 109 & 7.6M & 187 & 110 & 3.7M \\
7 & 1 & 103 & 44 & 2.9M & 98 & 109 & 8.1M & 192 & 111 & 4.1M \\
7 & 2 & 113 & 44 & 3.0M & 108 & 109 & 8.4M & 197 & 112 & 4.4M \\
7 & 3 & 119 & 44 & 3.1M & 112 & 110 & 8.9M & 199 & 113 & 4.9M \\
7 & 4 & 126 & 44 & 3.2M & 119 & 110 & 9.2M & 204 & 113 & 5.2M \\
\midrule
8 & 0 & 97 & 45 & 3.1M & 93 & 109 & 8.6M & 193 & 111 & 4.1M \\
8 & 1 & 104 & 45 & 3.2M & 99 & 109 & 9.1M & 197 & 111 & 4.5M \\
8 & 2 & 113 & 45 & 3.3M & 107 & 109 & 9.3M & 201 & 112 & 4.8M \\
8 & 3 & 118 & 45 & 3.4M & 111 & 110 & 9.9M & 203 & 113 & 5.3M \\
8 & 4 & 125 & 45 & 3.5M & 118 & 110 & 10M & 207 & 113 & 5.7M \\
\bottomrule
\end{tabular}
\end{minipage}

\caption{Appendix results across domains for LLaMA-4-Maverick.}
\label{tab:appendix-llama-4-maverick-split}
\end{table*}

\subsection{Heterogeneous Agent--User Experiments}
\label{app:heterogeneous_results}

We additionally evaluate DIVERT in heterogeneous GPT-OSS-120B/Gemini-2.5-Flash setups, where the agent and user simulator are instantiated with different models in both directions.
Across both settings, DIVERT consistently improves failure discovery efficiency and task-level coverage.
These trends match the main results, suggesting that DIVERT's benefits extend to cross-model agent--simulator interactions.

\begin{table}[H]
\centering
\small
\setlength{\tabcolsep}{8pt}
\caption{Heterogeneous setup with Gemini-2.5-Flash as agent and GPT-OSS-120B as user simulator over airline domain.}
\begin{tabular}{ccccc}
\toprule
RO & Branches & Err/100K $\uparrow$ & Fail C. $\uparrow$ & T $\downarrow$ \\
\midrule
2 & 0 & 12.7 & 29 & 355K \\
2 & 2 & 13.0 & 34 & 670K \\
2 & 4 & 13.9 & 38 & 884K \\
\midrule
4 & 0 & 12.7 & 31 & 715K \\
4 & 2 & 12.9 & 35 & 1.0M \\
4 & 4 & 13.9 & 38 & 1.3M \\
\midrule
6 & 0 & 13.1 & 33 & 1.0M \\
6 & 2 & 13.1 & 36 & 1.4M \\
6 & 4 & 13.9 & 38 & 1.6M \\
\bottomrule
\end{tabular}
\label{tab:heterogeneous-gemini-agent-gpt-user-airline}
\end{table}

\begin{table}[H]
\centering
\small
\setlength{\tabcolsep}{8pt}
\caption{Heterogeneous setup with GPT-OSS-120B as agent and Gemini-2.5-Flash as user simulator over airline domain.}
\begin{tabular}{ccccc}
\toprule
RO & Branches & Err/100K $\uparrow$ & Fail C. $\uparrow$ & T $\downarrow$ \\
\midrule
2 & 0 & 16.9 & 29 & 295K \\
2 & 2 & 18.2 & 34 & 544K \\
2 & 4 & 19.2 & 38 & 781K \\
\midrule
4 & 0 & 16.8 & 31 & 566K \\
4 & 2 & 17.7 & 35 & 815K \\
4 & 4 & 18.3 & 38 & 1.1M \\
\midrule
6 & 0 & 17.2 & 33 & 841K \\
6 & 2 & 17.8 & 36 & 1.1M \\
6 & 4 & 18.2 & 38 & 1.4M \\
\bottomrule
\end{tabular}
\label{tab:heterogeneous-gpt-agent-gemini-user-airline}
\end{table}

\subsection{Diversity Selection Validation}
\label{app:diversity_validation}

We report the full per-domain results for the two diversity-selection analyses discussed in Section~\ref{sec:diversity_validation}.
In both analyses, lower similarity indicates greater divergence from the original user utterance or trajectory.

\paragraph{Candidate-level diversity.}
We first measure the similarity between each generated candidate response and the original user utterance at the selected junction.
Table~\ref{tab:diversity-candidate-validation} shows that the most dissimilar candidate is consistently separated from the second and third candidates across all domains.

\begin{table}[H]
\centering
\small
\caption{Candidate-level diversity validation. Values report similarity between generated candidate responses and the original user utterance. Lower values indicate more dissimilar candidates.}
\begin{tabular}{lcccc}
\toprule
Domain & Most Dis. & 2nd Dis. & 3rd Dis. & N \\
\midrule
Airline & 0.721 & 0.772 & 0.816 & 1500 \\
Retail & 0.706 & 0.762 & 0.809 & 1500 \\
Telecom & 0.710 & 0.774 & 0.827 & 1500 \\
\midrule
All & 0.711 & 0.769 & 0.817 & 4500 \\
\bottomrule
\end{tabular}
\label{tab:diversity-candidate-validation}
\end{table}

\paragraph{Trajectory-level diversity.}
We next measure whether candidate-level diversity leads to downstream divergence by continuing each candidate from the same junction and comparing the resulting conversation suffix with the original suffix.
As shown in Table~\ref{tab:diversity-trajectory-validation}, the ordering is preserved downstream: candidates that are more dissimilar to the original utterance also lead to more divergent interaction trajectories.

\begin{table}[H]
\centering
\small
\caption{Trajectory-level diversity validation. Values report similarity between the resulting conversation suffix and the original suffix after branching from the same junction. Lower values indicate more divergent continuations.}
\begin{tabular}{lcccc}
\toprule
Domain & Most Dis. & 2nd Dis. & 3rd Dis. & N \\
\midrule
Airline & 0.741 & 0.797 & 0.851 & 400 \\
Retail & 0.734 & 0.789 & 0.841 & 400 \\
Telecom & 0.752 & 0.810 & 0.857 & 400 \\
\midrule
All & 0.743 & 0.799 & 0.849 & 1200 \\
\bottomrule
\end{tabular}
\label{tab:diversity-trajectory-validation}
\end{table}

\newpage
\section{Cost Breakdown and Token Accounting}
\label{app:cost_breakdown}

We provide detailed token usage statistics comparing full rollouts to DIVERT evaluation.

\paragraph{Per-Trajectory Token Profile.}

We conducted a numerical analysis of token consumption for standard full rollouts on the Airline domain.
On average, a single trajectory consumes 5,763.52 total tokens.
Of these, 4,068.86 (70.6\%) are generated by the agent, while 1,694.66 (29.4\%) are generated by the user simulator.

Under our method, the average number of agent tokens per trajectory drops to 3,272.96, corresponding to a reduction of 795.90 tokens (19.6\%) compared to full rollouts.
In addition, evaluation-side (user and framework) tokens decrease by 563.56 tokens per trajectory on average, reflecting reduced regeneration of shared prefixes and avoided repeated simulation steps.

Overall, this yields a \textbf{total average reduction of 795.90 agent tokens and 563.56 evaluation tokens} per trajectory prior to accounting for DIVERT overhead.

\subsection{Shared Prefix Analysis and KV-Cache Implications}
\label{app:shared_prefix_analysis}

We do not directly benchmark system-level KV-cache reuse, latency, or memory behavior, since these depend on the serving stack, batching policy, cache management, and hardware constraints.
Instead, we measure exact shared-prefix overlap as a simple proxy for whether such reuse is structurally available.
This analysis compares independent regular rollouts against DIVERT branches, where executions resume from identical cached prefixes before diverging.

\begin{table}[H]
\centering
\small
\caption{Average fraction of tokens belonging to exact shared prefixes across executions. Higher values indicate more computation that is identical before trajectories diverge, and therefore greater potential for KV-cache reuse under compatible serving systems.}
\begin{tabular}{lcc}
\toprule
Domain & Regular Prefix (\%) & Branch Prefix (\%) \\
\midrule
Airline & 0.59 & 34.80 \\
Retail & 0.51 & 42.50 \\
Telecom & 0.54 & 58.35 \\
\bottomrule
\end{tabular}
\label{tab:shared_prefix_kv}
\end{table}

As shown in Table~\ref{tab:shared_prefix_kv}, standard independent rollouts share almost no exact prefix tokens across executions (roughly $0.5\%$--$0.6\%$), even when they remain semantically similar.
In contrast, DIVERT substantially increases exact shared-prefix length, reaching $34.80\%$ on Airline, $42.50\%$ on Retail, and $58.35\%$ on Telecom.
These results should not be read as measured KV-cache speedups.
Rather, they show that DIVERT creates the exact-prefix structure required for potential cache reuse, whereas near-identical but non-exact rollout prefixes typically cannot share KV-cache.

\subsection{DIVERT Overhead}
\label{divert_overhead}

Each branch introduces additional framework overhead due to junction selection and divergent user generation.

The average token cost per branch consists of:

\begin{itemize}
    \item \textbf{Junction selection:} 103.48 tokens.
    \item \textbf{Candidate generation:} 108.68 tokens per candidate.
\end{itemize}

Since we generate $k=3$ candidates per junction, the total candidate generation cost is:
\[
3 \times 108.68 = 326.04 \text{ tokens}.
\]

Thus, the total average overhead per branch is:
\[
103.48 + 326.04 = 429.52 \text{ tokens}.
\]

\paragraph{Net Token Savings Per Trajectory.}

Beyond reducing agent tokens, branching also reduces evaluation-side (user/framework) tokens by:

\[
Y = 563.56 \text{ tokens}.
\]

Therefore, the total tokens saved per branch relative to an equivalent full-rollout continuation is:

\[
X + Y = 795.90 + 563.56 = 1,359.46.
\]

After subtracting the 429.52-token branching overhead, the net token savings per branch is:

\[
1,359.46 - 429.52 = 929.94 \text{ tokens}.
\]

This demonstrates that even after accounting for framework overhead, branching yields a substantial net reduction in total evaluation token usage, in addition to the reduction in agent tokens—which constitute the dominant and economically most significant component of the cost.


\subsection{Relative Cost of DIVERT Framework Overhead}
\label{app:divert_relative_cost}

We now estimate the monetary contribution of DIVERT framework overhead relative to the total evaluation cost under a realistic frontier-agent setting.

Let:
\[
\bar{T}_{\text{agent}} = 4{,}068.86
\]
denote the average agent tokens per trajectory (Appendix~\ref{app:cost_breakdown}),
and let
\[
\bar{T}_{\text{DIVERT}} = 429.52
\]
denote the average tokens required to create a branch with divert (Appendix \ref{divert_overhead})

Let $p_{\text{agent}}$ be the output-token price of a frontier agent (e.g., \$10--\$25 per 1M tokens),
and let $p_{\text{eval}}$ be the output-token price of GPT-OSS-120B (\$0.19 per 1M tokens; Table~\ref{tab:api_costs}).

The relative monetary contribution of DIVERT overhead per trajectory is therefore:

\[
\frac{
\bar{T}_{\text{DIVERT}} \cdot p_{\text{eval}}
}{
\bar{T}_{\text{agent}} \cdot p_{\text{agent}}
}.
\]

Using a representative frontier price of \$10 per 1M output tokens:

\[
\frac{
429.52 \times 0.19
}{
4{,}068.86 \times 10
}
=
\frac{81.61}{40{,}688.6}
\approx 0.0020
=
0.20\%.
\]

Under higher-tier pricing (e.g., \$25 per 1M output tokens), this fraction drops further:

\[
\frac{
429.52 \times 0.19
}{
4{,}068.86 \times 25
}
\approx 0.08\%.
\]

Thus, when evaluating a SOTA frontier agent, DIVERT framework overhead contributes well below 1\% of the total monetary evaluation cost.

Importantly, this estimate is conservative: GPT-OSS-120B can be hosted locally as an open-weight model, further reducing evaluation-side cost in practice.

These calculations confirm that the dominant cost driver is agent token generation, while junction selection and divergent user generation contribute only a marginal fraction of the total expense.

\paragraph{Why We Optimize Agent Tokens.}

In realistic deployment settings, the agent under evaluation is typically a \emph{top-tier production model}. Organizations deploying customer-facing agents rarely choose mid-tier backbones—they select the strongest available frontier models to maximize reliability, reasoning quality, and policy adherence. These models are not only more expensive per token, but also tend to generate substantially longer trajectories due to extended reasoning, reflection traces, tool-calling scaffolds, and multi-step interactions.

As a result, agent tokens are expensive along \emph{two dimensions}:
(i) higher per-token API pricing, and
(ii) larger per-trajectory token volume.

Table~\ref{tab:api_costs} shows representative pricing for several flagship API-based frontier models as of the paper submission date (14/02/25). Premium production models such as Claude Opus, OpenAI GPT-series flagship models, and Google Gemini Pro-tier models typically charge multiple dollars per million input tokens and up to \$10--\$25 per million output tokens. Since agent trajectories are output-heavy and frequently include long reasoning traces, the effective evaluation cost scales primarily with output generation.

\begin{table}[H]
\centering
\small
\caption{Representative frontier API pricing (per 1M tokens). Prices correspond to publicly listed flagship production tiers as of the paper submission date (14/02/25).}
\begin{tabular}{lccc}
\toprule
Model & Provider & Input (\$/1M) & Output (\$/1M) \\
\midrule
\multicolumn{4}{c}{\textit{Agent Models (Frontier APIs)}} \\
\midrule
Claude Opus  & Anthropic & 5.00 & 25.00 \\
GPT-5  & OpenAI & 1.25 & 10.00 \\
Gemini 3 Pro & Google & 2.00 & 12.00 \\
\midrule
\multicolumn{4}{c}{\textit{Evaluation Model}} \\
\midrule
GPT-OSS 120B & OpenAI & 0.039 & 0.19 \\
\bottomrule
\end{tabular}
\label{tab:api_costs}
\end{table}

\noindent\textbf{Concrete cost estimate ($\tau$-bench).}
In our setup, $\tau$-bench contains $278$ tasks 
($114$ Airline, $114$ Retail, $50$ Telecom).
With an average trajectory length of 4,068.86 agent generation tokens (Appendix~\ref{app:cost_breakdown}), a single full pass over all tasks requires
\[
278 \times 4{,}068.86 \approx 1.13\text{M}
\]
output tokens.

Under a common reliability setting of $10$ rollouts per task,
this scales to
\[
278 \times 10 \times 4{,}068.86
\approx 11.31\text{M}
\]
output tokens in total.

Under a branch-augmented setting of $10$ rollouts + $12$ branches per task, where each branch consumes on average 3,272.96 agent tokens (Appendix~\ref{app:cost_breakdown}), the total becomes
\[
278 \times \big(10 \times 4{,}068.86 + 12 \times 3{,}272.96 \big)
\approx 22.23\text{M}
\]
agent output tokens.

Even conservatively counting only output tokens (ignoring input-token charges),
the resulting API cost is:

\begin{table}[H]
\centering
\small
\caption{
Estimated $\tau$-bench inference cost assuming
10 rollouts + 12 branches per task and counting output tokens only
(22.23M generated tokens).
}
\begin{tabular}{lc}
\toprule
Model & Output-Only Cost (\$) \\
\midrule
Claude Opus (25 \$/1M) & 555.7 \\
GPT-5 (10 \$/1M) & 222.3 \\
Gemini 3 Pro (12 \$/1M) & 266.8 \\
\bottomrule
\end{tabular}
\label{tab:taubench_cost_estimate}
\end{table}

\noindent
This highlights why we did not run large-scale sweeps over closed frontier APIs: comprehensive evaluation quickly becomes financially heavy.
More importantly, it also motivates our optimization target: reducing \emph{agent} token usage yields the largest practical savings precisely in the realistic setting where the deployed agent is a costly frontier API model. In contrast, the additional overhead introduced by the evaluation framework (junction selection, user simulation, replay bookkeeping) can be implemented with cheaper or open-weight models and is not the dominant cost driver.

Therefore, the economically meaningful quantity to optimize is the number of \textbf{agent tokens generated}. \emph{Errors per 100K Agent Tokens} directly reflects the real-world cost of assessing a production-grade agent. Any reduction in redundant prefix regeneration or unnecessary full rollouts translates into immediate monetary savings when evaluating costly frontier systems.

For transparency, we additionally report full framework overhead, including user-model and junction-selection tokens. However, agent token efficiency remains the primary optimization target, as it captures the dominant and practically relevant cost driver in real-world agent evaluation pipelines.

\section{User Intent Preservation Analysis}
\label{app:user_intent_alignment}

For transparency and completeness, we additionally verify that branched user messages remain aligned with the original task objective by running an automatic intent-alignment analysis comparing original simulator messages and branched messages against the task purpose definition.

\paragraph{Scope.}
Among the $\tau$-bench domains, only the Airline domain provides an explicit per-task purpose field in the original benchmark specification. These tasks are also the most complex and constraint-heavy in the suite. We therefore use Airline as a strict proxy domain for intent preservation and task-objective alignment analysis.

\paragraph{Setup.}
For each evaluated instance, we extract the task purpose from the benchmark task specification and pair it with both the original user message and the branched user message generated at the same junction. Each message is evaluated independently with an LLM-as-a-judge using \textit{GPT-OSS-120B} at temperature 0, a strong open-weight instruction-tuned model. The judge receives the task purpose and a single user message, and determines whether the message reasonably attempts to achieve the same core goal.

\paragraph{Judge Prompt.}
The judge is instructed to focus strictly on intent and goal direction, not wording, quality, or strategy. The exact system prompt is:

\begin{prompt}[label={box:user_intent_prompt}]{Blue}{Intent Alignment Judge Prompt}

\small
\textbf{System Prompt}

You are an expert evaluator assessing whether a user's message preserves the intent of a given scenario purpose. Your task is to decide whether the user message is reasonably attempting to achieve the same core goal described in the purpose.

Focus only on core intent and goal direction, not wording, style, or timing. Treat messages as aligned if they pursue the same underlying objective, even if phrased differently. Do not judge quality or completeness. Mark NO only if the goal clearly changes, is contradicted, or becomes unrelated.

Respond with ONLY YES or NO, then give a brief explanation.

\vspace{0.5em}
\textbf{User Prompt Template}

Task Purpose: \texttt{<purpose>} \\
User Message: \texttt{<message>} \\
Does this user message attempt to satisfy the task purpose?

\end{prompt}

\paragraph{Procedure.}
We run this check over all collected branching points in the Airline domain. Original and branched messages are judged separately, producing intent-preserved vs.\ intent-missed labels. The analysis is performed post-hoc and does not affect branching decisions or evaluation cost. In addition, we manually inspect a sample of judge decisions to validate correctness and calibration.

\paragraph{Results.}
On the Airline domain (700 evaluated instances of the 50 airline tasks on repeated runs), the original simulator messages are judged to miss the task intent in 207 cases (28.12\%), while branched messages miss intent in 186 cases (25.27\%). This indicates that intent-directed branching does not degrade alignment and in practice slightly improves it, correcting a subset of cases where the original simulator message partially drifted from the task objective.

\paragraph{Conclusion.}
These results support that divergent user generation in DIVERT maintains task-purpose alignment under a strict automatic judge, with sampled human inspection yielding consistent judgments. Intent preservation therefore holds while still enabling broader behavioral exploration.

\section{Reproducibility Details}
\label{app:reproducibility}

\subsection{Randomness and Seeds}

All experiments were conducted with a fixed random seed of \textit{42}.
The same seed was used for agent rollouts, user generation, and branching logic to ensure determinism across repeated runs.

\subsection{Models and Decoding Configuration}

Unless otherwise stated, we use the same model backbone for both the acting agent and the user simulator.
Our Decoding and models configurations matching the original $\tau$-bench setup:
\begin{itemize}
    \item Agent temperature: \textit{0.0}
    \item User temperature: \textit{0.7}
    \item Maximum simulation steps per trajectory: \textit{100} 
    \item Maximum allowed errors per trajectory: \textit{10}
    \item LLM maximum retries per call: \textit{3}
\end{itemize}

No failed inference calls or missing runs were observed due to LLM errors.

\subsection{Execution Limits}

Trajectory execution terminates upon:
\begin{itemize}
    \item Successful task completion,
    \item Reaching the maximum step limit (100 turns), or
    \item Exceeding the maximum error threshold (10 errors).
\end{itemize}

\subsection{Code Availability}

The code will be made publicly available upon publication at \href{https://github.com/IBM/DIVERT_Efficient_Agent_Evaluation_via_Diversity_Guided_User_Simulation}{GitHub repository}.

\end{document}